\documentclass[manuscript,screen, nonacm]{acmart}

\AtBeginDocument{%
  \providecommand\BibTeX{{%
    \normalfont B\kern-0.5em{\scshape i\kern-0.25em b}\kern-0.8em\TeX}}}

\usepackage{xcolor}
\usepackage{tabularx}
\usepackage{multirow}
\usepackage[textsize=tiny]{todonotes}

\usepackage{calc}
\usepackage{tikz}
\usetikzlibrary{tikzmark}

\usepackage{ulem}
\usepackage{lipsum}
\usepackage{pdfcomment}
\usepackage{adjustbox}
\usepackage{changepage}

\usepackage[most]{tcolorbox}

\usepackage{framed}

\usepackage{mdframed}
\usepackage{graphicx}
\usepackage[inline]{enumitem}

\usepackage[utf8]{inputenc}
\usepackage{tcolorbox}
\tcbuselibrary{skins}

\newtcolorbox{fancytextbox}[1]{
  enhanced,
  colframe=gray,
  colback=white,
  size=small,
  boxrule=1pt,
  arc=2mm,
  fonttitle=\bfseries,
  title=#1,
  left=1mm,
  right=1mm,
  top=1mm,
  bottom=1mm,
  boxsep=1mm,
  width=0.9\linewidth,
}

\newtcolorbox{conclusionbox}{
  enhanced, colback=gray!20, colframe=blue, fonttitle=\bfseries, 
  left=6pt, right=6pt, top=6pt, bottom=6pt, 
  sharp corners, rounded corners=southwest, arc=3mm,
  title=Conclusion:
}

\newtcolorbox{applebox}[1]{
  enhanced,
  colback=white,
  colframe=gray!40,
  fonttitle=\bfseries,
  coltitle=black,
  colbacktitle=white,
  left=6pt,
  right=6pt,
  top=2pt,
  bottom=2pt,
  boxsep=5pt,
  boxrule=1.5pt,
  arc=1.5mm,
  title=#1, %
  borderline={0pt}{0pt}{white},
  attach boxed title to top left={yshift=-2mm, xshift=3mm},
  boxed title style={sharp corners, boxrule=0pt, colback=white, frame hidden, left=2pt, right=2pt},
}

\begin{document}

\definecolor{BLUE}{RGB}{0,0,255}
\definecolor{RED}{RGB}{255,0,0}
\definecolor{BLACK}{RGB}{0,0,0}
\definecolor{GreyBlue}{RGB}{132,151,176}
\definecolor{GreyPink}{RGB}{228, 173, 181}
\definecolor{GreyGreen}{RGB}{135,174,136}
\definecolor{YGrey}{RGB}{175,171,171}

\definecolor{GreyBlue2}{RGB}{132,151,255}
\definecolor{GreyPink2}{RGB}{255,173,181}
\definecolor{GreyGreen2}{RGB}{135,222,136}
\title{Harnessing the Power of LLMs in Practice: A Survey on ChatGPT and Beyond}

\author{Jingfeng Yang}
\authornote{These authors contributed equally.}
\email{jingfengyangpku@gmail.com}
\affiliation{%
  \institution{Amazon}
  \country{USA}
}
\author{Hongye Jin}
\email{jhy0410@tamu.edu}
\affiliation{%
  \institution{Department of Computer Science and Engineering, Texas A\&M University}
  \country{USA}
}
\authornotemark[1]
\author{Ruixiang Tang}
\email{rt39@rice.edu}
\affiliation{%
  \institution{Department of Computer Science, Rice University}
  \country{USA}
}
\authornotemark[1]
\author{Xiaotian Han}
\email{han@tamu.edu}
\affiliation{%
  \institution{Department of Computer Science and Engineering, Texas A\&M University}
  \country{USA}
}
\authornotemark[1]
\author{Qizhang Feng}
\email{qf31@tamu.edu}
\affiliation{%
  \institution{Department of Computer Science and Engineering, Texas A\&M University}
  \country{USA}
}
\authornotemark[1]
\author{Haoming Jiang}
\email{jhaoming@amazon.com}
\affiliation{%
  \institution{Amazon}
  \country{USA}
}
\author{Bing Yin}
\email{alexbyin@amazon.com}
\affiliation{%
  \institution{Amazon}
  \country{USA}
}
\author{Xia Hu}
\email{xia.hu@rice.edu}
\affiliation{%
  \institution{Department of Computer Science, Rice University}
  \country{USA}
}

\begin{abstract}

This paper presents a comprehensive and practical guide for practitioners and end-users working with Large Language Models (LLMs) in their downstream natural language processing (NLP) tasks. We provide discussions and insights into the usage of LLMs from the perspectives of models, data, and downstream tasks. Firstly, we offer an introduction and brief summary of current GPT- and BERT-style LLMs. Then, we discuss the influence of pre-training data, training data, and test data. 
Most importantly, we provide a detailed discussion about the use and non-use cases of large language models for various natural language processing tasks, such as knowledge-intensive tasks, traditional natural language understanding tasks, natural language generation tasks, emergent abilities, and considerations for specific tasks.
We present various use cases and non-use cases to illustrate the practical applications and limitations of LLMs in real-world scenarios. 
We also try to understand the importance of data and the specific challenges associated with each NLP task. 
Furthermore, we explore the impact of spurious biases on LLMs and delve into other essential considerations, such as efficiency, cost, and latency, 
to ensure a comprehensive understanding of deploying LLMs in practice. This comprehensive guide aims to provide researchers and practitioners with valuable insights and best practices for working with LLMs, thereby enabling the successful implementation of these models in a wide range of NLP tasks. A curated list of practical guide resources of LLMs, regularly updated, can be found at \url{https://github.com/Mooler0410/LLMsPracticalGuide}.

\end{abstract}



\maketitle

\section{Introduction}\label{sec:intro}
In recent years, the rapid development of Large language Models has been revolutionizing the field of natural language processing~\cite{zhao2023survey, zhou2023comprehensive, bommasani2021opportunities}. These powerful models have shown great potential in addressing a variety of NLP tasks, ranging from natural language understanding~(NLU) to generation tasks, even paving the way to Artificial General Intelligence (AGI). However, utilizing these models effectively and efficiently requires a practical understanding of their capabilities and limitations, as well as the data and tasks involved in NLP. 

To provide a guide for partitioners and end-users, this work focuses on the practical aspects of working with LLMs in downstream NLP tasks. 
This guide aims to provide practical advice on why or why not to choose LLMs for a given task, as well as guidance on how to select the most suitable LLM, taking into account factors such as model sizes, computational requirements, and the availability of domain-specific pre-trained models.
This work offers a thorough understanding of LLMs from a practical perspective, therefore, empowers practitioners and end-users with the practical knowledge needed to successfully leverage the power of LLMs for their own NLP tasks.

Our work is structured as follows. First, our work offers a brief introduction to LLMs by discussing the most important models, such as GPT-style and BERT-style architectures. 
Then, we delve into the critical factors that influence model performance from the data perspective, including pre-training data, training/tuning data, and test data. 
Last and most importantly, we dive deep into various concrete NLP tasks, offering insights into the applicability of LLMs for knowledge-intensive tasks, traditional NLU tasks, and generation tasks, along with the emergent abilities that these models possess and challenging real-world scenarios. We provide detailed examples to highlight both the successful use cases and the limitations of LLMs in practice.

To analyze the abilities of large language models, we compare them with fine-tuned models. 
As of present, there is no universally recognized definition for LLMs and fine-tuned models. With consideration to practical utility, in our article, the definitions of them are proposed as: LLMs are huge language models pretrained on large amounts of datasets without tuning on data for specific tasks; fine-tuned models are typically smaller language models which are also pretrained and then further tuned on a smaller, task-specific dataset to optimize their performance on that task\footnote{From a practical standpoint, we consider models with less than 20B parameters to be fine-tuned models. While it's possible to fine-tune even larger models like PlaM (540B), in reality, it can be quite challenging, particularly for academic research labs and small teams. Fine-tuning a model with 3B parameters can still be a daunting task for many individuals or organizations.}. 

This work summarizes the following main practical guides for using LLMs:
\begin{itemize}
    \item \textbf{Natural language understanding.} Employ the exceptional generalization ability of LLMs when facing out-of-distribution data or with very few training data. 
    \item \textbf{Natural language generation.} Utilize LLMs' capabilities to create coherent, contextually relevant, and high-quality text for various applications.
    \item \textbf{Knowledge-intensive tasks.} Leverage the extensive knowledge stored in LLMs for tasks requiring domain-specific expertise or general world knowledge.
    \item \textbf{Reasoning ability.} Understand and harness the reasoning capabilities of LLMs to improve decision-making and problem-solving in various contexts.
\end{itemize}

\section{Practical Guide for Models}
\begin{figure}[tp]
  \begin{adjustwidth}{-0.0cm}{}
  \centering
    \includegraphics[width=1.00\textwidth]{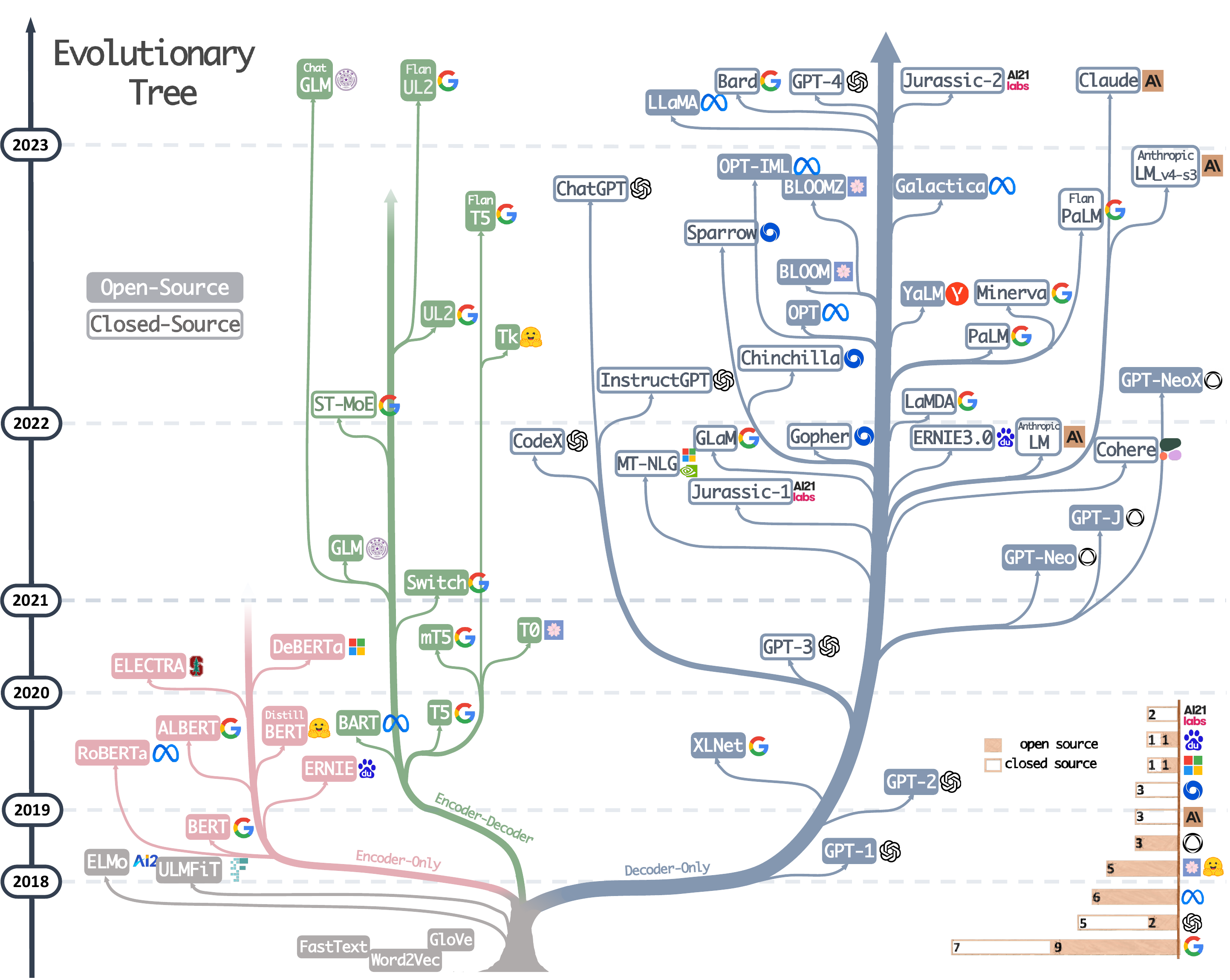}
   \end{adjustwidth} 
   \caption{The evolutionary tree of modern LLMs traces the development of language models in recent years and highlights some of the most well-known models. Models on the same branch have closer relationships. Transformer-based models are shown in non-\textcolor{YGrey}{grey} colors: decoder-only models in the \textcolor{GreyBlue2}{blue} branch, encoder-only models in the \textcolor{GreyPink2}{pink} branch, and encoder-decoder models in the \textcolor{GreyGreen2}{green} branch. The vertical position of the models on the timeline represents their release dates. Open-source models are represented by solid squares, while closed-source models are represented by hollow ones. The stacked bar plot in the bottom right corner shows the number of models from various companies and institutions.}\label{fig:tree}
\end{figure}

This section provides a brief introduction to state-of-the-art LLMs. These models differ in their training strategies, model architectures, and use cases. To provide a clearer understanding of the LLM landscape, we categorize them into two types: encoder-decoder or encoder-only language models and decoder-only language models. In Figure~\ref{fig:tree}, we show the detailed evolution process of language models. From the evolutionary tree, we make the following interesting observations: 
\begin{enumerate}[label=\alph*)]
    \item Decoder-only models have been gradually dominating the development of LLMs. At the early stage of LLMs development, \textcolor{GreyBlue}{decoder-only} models were not as popular as \textcolor{GreyPink}{encoder-only} and \textcolor{GreyGreen}{encoder-decoder} models. However, after 2021, with the introduction of game-changing LLMs - GPT-3, decoder-only models experienced a significant boom. Meanwhile, after the initial explosive growth brought about by BERT, encoder-only models gradually began to fade away.

    \item OpenAI consistently maintains its leadership position in LLM, both currently and potentially in the future. Other companies and institutions are struggling to catch up with OpenAI in developing models comparable to GPT-3 and the current GPT-4. This leadership position may be attributed to OpenAI's steadfast commitment to its technical path, even when it was not widely acknowledged initially.
    
    \item Meta contributes significantly to open-source LLMs and promotes research of LLMs. When considering contributions to the open-source community, particularly those related to LLMs, Meta stands out as one of the most generous commercial companies, as all the LLMs developed by Meta are open-sourced.

    \item LLMs exhibit a tendency towards closed-sourcing. In the early stages of LLM development (before 2020), the majority of models were open-sourced. However, with the introduction of GPT-3, companies have increasingly opted to close-source their models, such as PaLM, LaMDA, and GPT-4. Consequently, it has become more difficult for academic researchers to conduct experiments on LLM training. As a result, API-based research could become the predominant method in the academic community.

    \item Encoder-decoder models remain promising, as this type of architecture is still being actively explored, and most of them are open-sourced. Google has made substantial contributions to open-source encoder-decoder architectures. However, the flexibility and versatility of decoder-only models seem to make Google's insistence on this direction less promising.

\end{enumerate}
We also briefly summarize the characteristics and the representative LLMs of each type in Table~\ref{tab:llms}.

\begin{table}[tp]
\centering
\caption{Summary of Large Language Models.}
\label{tab:llms}
\resizebox{1\textwidth}{!}{
    \begin{tabular}{c|rl|c}
    \toprule
    \multicolumn{1}{c|}{}                       &\multicolumn{2}{c|}{\textbf{Characteristic}}        & \multicolumn{1}{c}{\textbf{LLMs}}     \\ \midrule
    \multirow{5}{*}{Encoder-Decoder or Encoder-only}     &              &                                    & \multirow{4}{6cm}{ELMo~\cite{peters2018deep}, BERT~\cite{devlin2018bert}, RoBERTa~\cite{liu2019roberta}, DistilBERT~\cite{sanh2019distilbert}, BioBERT~\cite{lee2020biobert}, XLM~\cite{lample2019cross}, Xlnet~\cite{yang2019xlnet}, ALBERT ~\cite{lan2019albert}, ELECTRA~\cite{clark2020electra}, T5~\cite{2020t5}, GLM~\cite{zeng2022glm}, XLM-E~\cite{chi2021xlm}, ST-MoE~\cite{zoph2202st}, AlexaTM~\cite{soltan2022alexatm}} \\ 
                                                &Training: &Masked Language Models         & \\
                                                &Model type:   &Discriminative                         & \\
    (BERT-style)                                 &Pretrain task:&Predict masked words                   & \\
                                                &              &                                       & \\
    \midrule
    \multirow{5}{*}{Decoder-only}      &              &                                       & \multirow{4}{6cm}{GPT-3~\cite{brown2020language}, OPT~\cite{zhang2022opt}. PaLM~\cite{chowdhery2022palm}, BLOOM~\cite{scao2022bloom}, MT-NLG~\cite{smith2022using},  GLaM~\cite{du2022glam},Gopher~\cite{rae2021scaling}, chinchilla~\cite{hoffmann2022training}, LaMDA~\cite{thoppilan2022lamda}, GPT-J~\cite{mesh-transformer-jax}, LLaMA~\cite{meta2023llama},  GPT-4~\cite{openai2023gpt4}, BloombergGPT~\cite{wu2023bloomberggpt}} \\
                                                         &Training  & Autoregressive Language Models                         & \\
                                                         &Model type:   &Generative                             & \\
    (GPT-style)                                           &Pretrain task:&Predict next word                      & \\
                                                         &              &                                       & \\
    \bottomrule
    \end{tabular}
}
\end{table}

\subsection{BERT-style Language Models: Encoder-Decoder or Encoder-only}

As natural language data is readily available and unsupervised training paradigms have been proposed to better utilize extremely large datasets, this motivates the unsupervised learning of natural language. One common approach is to predict masked words in a sentence while considering the surrounding context. This training paradigm is known as the Masked Language Model. This type of training allows the model to develop a deeper understanding of the relationships between words and the context in which they are used.  These models are trained on a large corpus of texts using techniques such as the Transformer architecture and have achieved state-of-the-art results in many NLP tasks, such as sentiment analysis and named entity recognition. Notable examples of Masked Language Models include BERT~\cite{devlin2018bert}, RoBERTa~\cite{liu2019roberta}, and T5~\cite{2020t5}. MLMs have become an important tool in the field of natural language processing due to their success in a wide range of tasks.

\subsection{GPT-style Language Models: Decoder-only}

Although language models are typically task-agnostic in architecture, these methods require fine-tuning on datasets of the specific downstream task. Researchers found that scaling up language models significantly improves the few-shot, even zero-shot performance~\citep{brown2020language}. The most successful models for better few-shot and zero-show performance are Autoregressive Language Models, which are trained by generating the next word in a sequence given the preceding words. These models have been widely used for downstream tasks such as text generation and question answering. Examples of Autoregressive Language Models include GPT-3~\cite{brown2020language}, OPT~\cite{zhang2022opt}, PaLM~\cite{chowdhery2022palm}, and BLOOM~\cite{scao2022bloom}. The game changer, GPT-3, for the first time, demonstrated reasonable few-/zero-shot performance via prompting and in-context learning,  
thus showing the superiority of autoregressive language models. There are also models such as CodeX~\cite{codex} that are optimized for specific tasks such as code generation, BloombergGPT~\cite{wu2023bloomberggpt} for the financial domain. The recent breakthrough is ChatGPT, which refines GPT-3 specifically for conversational tasks, resulting in more interactive, coherent, and context-aware conversational for various real-world applications.

\section{Practical Guide for Data}

In this section, we'll be discussing the critical role that data plays in selecting appropriate models for downstream tasks. The impact of data on the models' effectiveness starts during the pre-training stage and continues through to the training and inference stages.

\begin{applebox}{Remark 1}
\begin{enumerate}[leftmargin=0.4cm]

\item LLMs generalize better than fine-tuned models in downstream tasks facing out-of-distribution data, such as  adversarial examples and domain shifts.
\item LLMs are preferable to fine-tuned models when working with limited annotated data, and both can be reasonable choices when abundant annotated data is available, depending on specific task requirements.
\item It's advisable to choose models pre-trained on fields of data that are similar to downstream tasks.

\end{enumerate}
\end{applebox}

\subsection{Pretraining data}

Pre-training data plays a pivotal role in the development of large language models. As the foundation of remarkable capabilities \cite{alajrami2022does, kaplan2020scaling} of LLMs, the quality, quantitative, and diversity of pre-training data influence the performance of LLMs significantly~\cite{zha2023data}. The commonly used pretraining data consists of a myriad of text sources, including books, articles, and websites. The data is carefully curated to ensure a comprehensive representation of human knowledge, linguistic nuances, and cultural perspectives. The importance of pretraining data lies in its capacity to inform the language model with a rich understanding of word knowledge, grammar, syntax, and semantics, as well as the ability to recognize context and generate coherent responses.
The diversity of pretraining data also plays a crucial role in shaping the model's performance, and the selection of LLMs highly depends on the components of the pretraining data. For example, PaLM~\cite{chowdhery2022palm} and BLOOM~\cite{scao2022bloom} excel in multilingual tasks and machine translation with an abundance of multilingual pretraining data. Moreover, PaLM's performance in Question Answering tasks is enhanced by incorporating a considerable amount of social media conversations and Books corpus \cite{chowdhery2022palm}. Likewise,  code execution and code completion capabilities of GPT-3.5~(code-davinci-002) are amplified by the integration of code data in its pretraining dataset. In brief, when selecting LLMs for downstream tasks, it is advisable to choose the model pre-trained on a similar field of data.

\subsection{Finetuning data}
When deploying a model for downstream tasks, it is essential to consider three primary scenarios based on the availability of annotated data: zero, few, and abundant. In this section, we provide a succinct overview of the appropriate models to employ for each scenario.

\noindent\textbf{Zero annotated data}: 
In scenarios where annotated data is unavailable, utilizing LLMs in a zero-shot setting proves to be the most suitable approach. LLMs have been shown to outperform previous zero-shot methods \cite{yin2019benchmarking}. Additionally, the absence of a parameter update process ensures that catastrophic forgetting \cite{kirkpatrick2017overcoming} is avoided since the language model parameters remain unaltered.

\noindent\textbf{Few annotated data}: In this case, the few-shot examples are directly incorporated in the input prompt of LLMs, which is named as in-context learning, and these examples can effectively guide LLMs to generalize to the task. As reported in~\cite{brown2020language}, one-shot and few-shot performance make significant gains, even matching the performance of the SOTA fine-tuned open-domain models. And LLMs' zero/few-shot ability can be improved further by scaling~\cite{brown2020language}.  %
 Alternatively, some few-shot learning methods are invented to enhance fine-tuned models, such as meta-learning \cite{lee2022meta} or transfer learning \cite{ruder2019transfer}. However, performance might be inferior compared to using LLMs due to fine-tuned models' smaller scale and overfitting. 

\noindent\textbf{Abundant annotated data}: With a substantial amount of annotated data for a particular task available, both fine-tuned models and LLMs can be considered. In most cases, fine-tuning the model can fit the data pretty well. Although, LLMs can be used to meet some constraints such as privacy~\cite{tang2023does}.In this scenario, the choice between using a fine-tuned model or a LLM is task-specific and also depends on many factors, including desired performance, computational resources, and deployment constraints. 
 
In a brief summary: LLMs are more versatile w.r.t. the data availability, while fine-tuned models can be considered with abundant annotated data.

\subsection{Test data/user data }
\label{sec:test_data}

When deploying LLMs for downstream tasks, we often face challenges stemming from distributional differences between the test/user data
and that of the training data. These disparities may encompass domain shifts \cite{zhou2022domain}, out-of-distribution variations \cite{du2022shortcut}, or even adversarial examples \cite{qiu2022adversarial}. 
Such challenges significantly hinder fine-tuned modes' effectiveness in real-world applications. They fit into a specific distribution and have a poor ability to generalize to OOD data. However, LLMs perform quite well facing such scenarios because they do not have an explicit fitting process.
Moreover, recent advancements have further enhanced the ability of language models in this regard.
The Reinforcement Learning from Human Feedback (RLHF) method has notably enhanced LLMs' generalization capabilities~\cite{ouyang2022training}. For example, InstructGPT demonstrates proficiency in following various instructions for a wide range of tasks and occasionally complying with instructions in different languages, even though such instructions are scarce. 
Similarly, ChatGPT exhibits consistent advantages on most adversarial and out-of-distribution (OOD) classification and translation tasks \cite{wang2023robustness}. Its superiority in understanding dialogue-related texts led to an impressive performance on the DDXPlus dataset \cite{tchango2022ddxplus}, a medical diagnosis dataset designed for OOD evaluation.

\footnotetext{As we mention in Section \ref{sec:intro}, LLMs are pretrained on large and diverse datasets without fine-tuning, while fine-tuned models are typically pretrained on a large dataset and then further trained on a smaller, task-specific dataset to optimize their performance on that task.}

\section{Practical Guide for NLP Tasks}
\begin{figure}[tp]
    \centering
    \includegraphics[width=\textwidth]{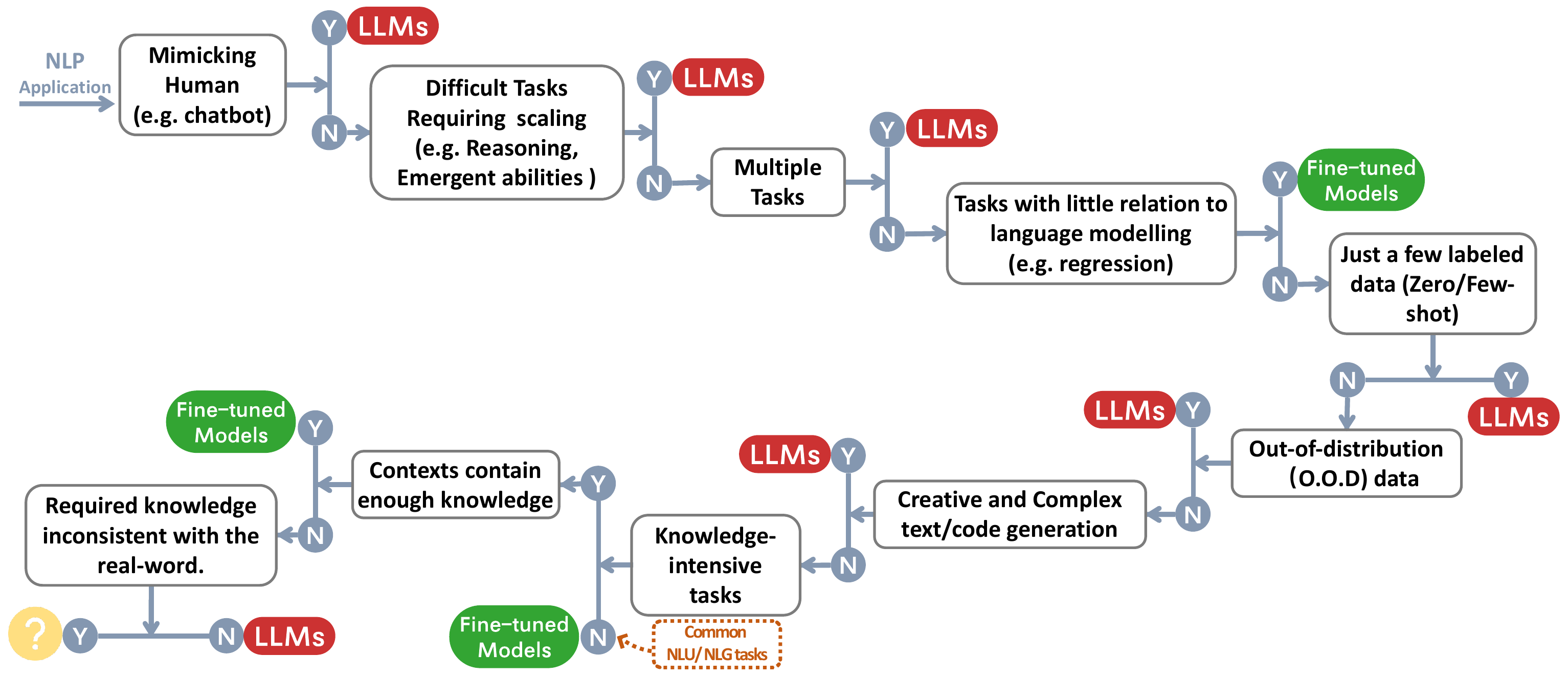}
    \caption{The decision flow for choosing LLMs or fine-tuned models~\protect\footnotemark for user's NLP applications. The decision flow helps users assess whether their downstream NLP applications at hand meet specific conditions and, based on that evaluation, determine whether LLMs or fine-tuned models are the most suitable choice for their applications. During the decision process in the figure, $\vcenter{\hbox{\includegraphics[scale=0.35]{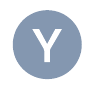}}}$ means meeting the condition, and $\vcenter{\hbox{\includegraphics[scale=0.35]{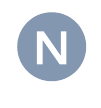}}}$ means not meeting the condition. The yellow circle for $\vcenter{\hbox{\includegraphics[scale=0.35]{./Y.pdf}}}$ of the last condition means there's no model working well on this kind of application.}\label{fig:decision} 
\end{figure}

In this section, we discuss in detail the use cases and no use cases for LLMs in various downstream NLP tasks and the corresponding model abilities. And in Figure~\ref{fig:decision}, we summarize all discussions into a decision flow. It can be a guide for a quick decision while facing a task.

\subsection{Traditional NLU tasks}

Traditional NLU tasks are some fundamental tasks in NLP including text classification, named entity recognition~(NER), entailment prediction, and so on. Many of them are designed to serve as intermediate steps in larger AI systems, such as NER for knowledge graph construction. 

\begin{applebox}{Remark 2}
    Fine-tuned models generally are a better choice than LLMs in traditional NLU tasks, but LLMs can provide help while requiring strong generalization ability.
\end{applebox}

\subsubsection{No use case}
In most natural language understanding tasks, such as tasks in GLUE\cite{wang2018glue} and SuperGLUE\cite{wang2019superglue}, fine-tuned models still have better performance, if such tasks come with rich well-annotated data and contain very few out-of-distribution examples on test sets. For different tasks and datasets, the gap between small fine-tuned models and LLMs varies. 

In text classification, on most datasets, LLMs perform slightly worse than fine-tuned models. %
For sentiment analysis, such as on IMDB~\cite{maas2011learning} and SST~\cite{socher2013recursive}, fine-tuned models and LLMs perform equally well. For toxicity detection, which is another iconic text classification task, the gap is much larger. All LLMs cannot perform well on this task, and on CivilComments~\cite{borkan2019nuanced} even the best one is only better than random guessing~\cite{liang2022holistic}. On the other hand, most popular fine-tuned models can obtain much better performance~\cite{duchene2023benchmark}. 
and the Perspective API~\footnote{https://perspectiveapi.com} is still one of the best for detecting toxicity. This API is powered by a multilingual BERT-based model, which is tuned on publicly available toxicity data 
and several smaller single-language CNNs distilled from this model.
This might be due to the fact that toxicity is defined by subtle nuances in linguistic expressions, and large language models are unable to accurately comprehend this task solely based on the provided input.  

The trend of performance gaps is similar in some other tasks. For natural language inference~(NLI) tasks, on most datasets, such as on RTE~\cite{wang2018glue} and SNLI~\cite{snli:emnlp2015}, fine-tuned models perform better than LLMs, while on some data such as CB~\cite{wang2019superglue}, LLMs have obtained comparable performance with fine-tuned models~\cite{chowdhery2022palm}. For question answering~(QA), on SQuADv2~\cite{rajpurkar2018know}, QuAC~\cite{choi2018quac} and many other datasets, fine-tuned models have superior performance, while on CoQA~\cite{reddy2019coqa}, LLMs perform as well as fine-tuned models~\cite{chowdhery2022palm}.  %

In information retrieval~(IR) tasks, LLMs are not widely exploited yet. One major reason is that IR tasks are fundamentally different from others. There's no natural way to transform the thousands of candidate texts into a few/zero-shot form which is required by LLMs. The existing evaluation results on MS MARCO(regular/TREC)~\cite{nguyen2016ms} show that methods based on fine-tuned models have better performance~\cite{liang2022holistic}. In this evaluation, the LLMs rank passages in an unorthodox way, which requires the LLMs to produce probabilities for passages one by one.

For some low-level intermediate tasks, which are not intended for regular users but rather for high level tasks, such as named entity recognition~(NER) and dependency parsing, there's not enough result coming from LLMs, because the most current evaluation of LLMs focuses on practical tasks. According to available evaluation results, for the NER task, CoNLL03~\cite{sang2003introduction} is still a challenge for LLMs~\cite{qin2023chatgpt}, where the performance of fine-tuned models is around as twice as LLMs. These intermediate tasks may vanish soon because LLMs can take over high-level tasks without the help of those intermediate tasks~(e.g. dependency parsing for coding tasks; NER for some text generation tasks).

In brief, for most traditional NLU tasks, a fine-tuned model is a better choice in terms of the performance on benchmark datasets and the computational cost. The scale of LLMs is usually $10\times$ or even $100\times$ larger than fine-tuned models. One possible cause for the inferior performance of LLMs on certain tasks can be the design of instructions/prompts. Transforming input from tasks like IR and sentence labeling into a few/zero-short instruction form is non-trivial. There may be better ways to adapt language models to traditional NLP tasks in the future. On the other hand, the upper limit of capabilities of fine-tuned models is not reached, and some methods like FLAN-tuning~\cite{longpre2023flan} can further boost the performance on NLU tasks. Another interesting finding is that on NLU tasks, after fine-tuning, masked language models, like T5\cite{raffel2020exploring}, are better than most auto-regressive language models at the same scale, while some recent results imply that this gap can be bridged by scaling\cite{chowdhery2022palm}.

\subsubsection{Use case}
However, there are still some NLU tasks suitable for LLMs.

One of the representative tasks is miscellaneous text classification~\cite{liang2022holistic}. In contrast to classic domain-specific text classification tasks such as sentiment analysis, miscellaneous text classification deals with a diverse range of topics and categories that may not have a clear or strong relationship with one another. It's closer to real-world cases and hard to be formatted for using fine-tuned models.
Another is the Adversarial NLI~(ANLI)\cite{nie2019adversarial}. It is a difficult dataset composed of adversarially mined natural language inference questions in three rounds (R1, R2, and R3). LLMs have shown superior performance on ANLI, especially on the R3 and R2. Both examples demonstrate the exceptional ability of LLMs to generalize well on out-of-distribution and sparsely annotated data in traditional NLP tasks, surpassing that of fine-tuned models. We've discussed this in the section above \ref{sec:test_data}.

\subsection{Generation tasks}
Natural Language Generation broadly encompasses two major categories of tasks, with the goal of creating coherent, meaningful, and contextually appropriate sequences of symbols. The first type focuses on converting input texts into new symbol sequences, as exemplified by tasks like paragraph summarization and machine translation. The second type, "open-ended" generation, aims to generate text or symbols from scratch to accurately match input descriptions such as crafting emails, composing news articles, creating fictional stories and writing code. 

\begin{applebox}{Remark 3}
    Due to their strong generation ability and creativity, LLMs show superiority at most generation tasks. 
\end{applebox}

\subsubsection{Use case} Generation tasks require models to have a comprehensive understanding of the input contents or requirements and a certain level of creativity. This is what LLMs excel at. 

For summarization tasks, although LLMs do not have an obvious advantage over fine-tuned models under traditional automatic evaluation metrics, such as ROUGE~\cite{lin2004rouge}, human evaluation results indicate that humans tend to prefer the results generated by LLMs~\cite{goyal2022news, zhang2023benchmarking} compared to that of fine-tuned models. For example, on CNN/DailyMail~\cite{nallapati2016abstractive} and XSUM~\cite{narayan2018don}, fine-tuned models like Brio~\cite{liu2022brio} and Pegasus~\cite{zhang2020pegasus} have much better performance than any LLMs w.r.t. ROUGE,  but LLMs like OPT~\cite{zhang2022opt} perform far better in human evaluation considering all aspects including faithfulness, coherence, and relevance~\cite{zhang2023benchmarking}. This demonstrates the superiority of LLMs in summarization tasks. On the other hand, it implies that current summarization benchmarks don't contain summaries with high quality or the automatic metrics are not proper for the evaluation of summarization.

In machine translation~(MT), LLMs can perform competent translation, although the average performance is slightly worse than some commercial translation tools~\cite{jiaochatgpt} considering some automatic metrics like BLEU\cite{papineni2002bleu}. LLMs are particularly good at translating some low-resource language texts to English texts, such as in the Romanian-English translation of WMT'16~\cite{bojar-etal-2016-findings}, zero-shot or few-shot LLMs can perform better than SOTA fine-tuned model\cite{chowdhery2022palm}. This is mainly due to the fact that English resources compose the main part of the pre-training data. BLOOM~\cite{scao2022bloom} is pre-trained on more multi-lingual data, leading to better translation quality in both rich-resource and low-resource translation. Another interesting finding is that BLOOM achieves good translation quality among Romance languages, even for translation from Galician, which is not included in the pre-training data. One reasonable explanation is that texts from some languages in the same language group can help the LLMs learn more from the similarity. 
If more multi-lingual texts can be added to the pre-training data, the translation capability may be improved further.

 Additionally, LLMs are highly skilled in open-ended generations. One example is that the news articles generated by LLMs are almost indistinguishable from real news articles by humans~\cite{brown2020language}.
 LLMs are remarkably adept at code synthesis as well. Either for text-code generation, such as HumanEval~\cite{chen2021evaluating} and MBPP~\cite{austin2021program}, or for code repairing, such as DeepFix~\cite{gupta2017deepfix}, LLMs can perform pretty well. GPT-4 can even pass 25\% problems in Leetcode, which are not trivial for most human coders~\cite{openai2023gpt4}. With training on more code data, the coding capability of LLMs can be improved further~\cite{chowdhery2022palm}. While performing well on such tasks, the codes generated by LLMs should be tested carefully to figure out any subtle bugs, which is one of the main challenges for applying LLMs in code synthesis.

\subsubsection{No use case} 
Fine-tuned models, such as DeltaLM+Zcode~\cite{yang-etal-2021-multilingual-machine}, still perform best on most rich-resource translation and extremely low-resource translation tasks. In rich resource machine translation, fine-tuned models slightly outperform LLMs~\cite{chowdhery2022palm, scao2022bloom}. And in extremely low-resource machine translation, such as English-Kazakh translation, fine-tuned models significantly perform better than LLMs.

\subsection{Knowledge-intensive tasks}
Knowledge-intensive NLP tasks refer to a category of tasks that have a strong reliance on background knowledge, domain-specific expertise, or general real-world knowledge. These tasks go beyond simple pattern recognition or syntax analysis. And they are highly dependent on memorization and proper utilization of knowledge about specific entities, events, and common sense of our real world.

\begin{applebox}{Remark 4}
    \begin{enumerate}[leftmargin=0.4cm]
        \item LLMs excel at knowledge-intensive tasks due to their massive real-world knowledge.
        \item LLMs struggle when the knowledge requirements do not match their learned knowledge, or when they face tasks that only require contextual knowledge, in which case fine-tuned models can work as well as LLMs.
    \end{enumerate}
\end{applebox}

\subsubsection{Use case}In general, with billions of training tokens and parameters,  LLMs have much more real-world knowledge than fine-tuned models.  

Closed-book question-answering tasks require the model to answer a given question about factual knowledge without any external information. It does require the memorization of real-world knowledge in the model. LLMs perform better on nearly all datasets, such as on NaturalQuestions~\cite{kwiatkowski2019natural}, WebQuestions~\cite{berant2013semantic}, and TriviaQA~\cite{joshi2017triviaqa}. On TriviaQA, even zero-shot LLMs is still much better~\cite{chowdhery2022palm}.

The massive multitask language understanding~(MMLU)~\cite{hendrycks2020measuring} is also highly knowledge-intensive. It contains multiple-choice questions spanning over 57 different subjects and requires general knowledge of the model. It's pretty challenging even for LLMs, although the newly released GPT-4~\cite{openai2023gpt4} outperforms existing models by a considerable margin in English with a satisfactory 86.5\% accuracy.

Also, some tasks in Big-bench\cite{srivastava2022beyond}, which are designed to probe LLMs and extrapolate their future capabilities, heavily relied on the memorization of real-world knowledge. In such tasks, the performance of some LLMs is better than the average level of humans, and even comparable to the best human performance. For example, the task \textit{Hindu\_knowledge} requires models to give facts about Hindu mythology, \textit{Periodic Elements} require the capability of predicting the element name from the periodic table and \textit{Physics} tests the physics knowledge of models by asking for the formula needed to solve a given physics problem.

\subsubsection{No use case}
There are some other tasks requiring knowledge different from that learned by LLMs. The required knowledge is not that learned by LLMs about the real world. In such tasks, LLMs are not notably superior.

Some tasks only require the model to capture the self-contained knowledge in the contexts. The knowledge in the contexts from the input is enough for the model to make predictions. For these tasks, small fine-tuned models can work pretty well. One such task is machine reading comprehension~(MRC). An MRC task provides several paragraphs and  requires the model to predict the answer to questions based on these paragraphs. We've discussed MRC in the previous section because it's also a traditional NLU task. 

Another scenario is that the knowledge within LLMs about real world is useless to the task, or even the required knowledge is counterfactual to the real world. As a result, the LLMs cannot work well on such tasks. In some cases, inconsistent knowledge may even make the LLMs worse than random guessing. For example, in Big-Bench, the Mnist ascii task requires the model to tell the digit represented by an ASCII art. The capability required by this task is nothing about real-world knowledge. Also, in the Inverse Scaling Phenomenon competition~\cite{mckenzie2022round2}, the task \textit{redefine math} redefines a common symbol and requires the model to choose between the original meaning and the meaning derived from the redefinition. What it requires contrasts to the LLMs' knowledge, thus LLMs even perform worse than random guessing.  

As an alternative to real-world knowledge in LLMs, access to extra knowledge is allowed, and models can thus get enough knowledge for a task via retrieval augmentation. The basic idea of retrieval augmentation is to add an extra information retrieval step prior to making predictions, in which, some useful texts related to the task will be retrieved from a large corpus. Then, the model will make predictions based on both the input contexts and the retrieved texts. With retrieved additional information, the closed-book task can become "open-book". In such a scenario, fine-tuned models are pretty good with much smaller sizes, because the required knowledge can be obtained by retrieving. For example, on NaturalQuestions~\cite{kwiatkowski2019natural}, with extra corpus, retrieval augmented models~\cite{kedia2022fie, izacard_few-shot_2022} are much better than any other methods.

\subsection{Abilities Regarding Scaling}

Scaling of LLMs~(e.g. parameters, training computation, etc.) can greatly empower pretrained language models. With the model scaling up, a model generally becomes more capable in a range of tasks. Reflected in some metrics, the performance shows a power-law relationship with the model scale. For example, the cross-entropy loss which is used to measure the performance for language modeling decreases linearly with the exponential increase in the model scale, which is also called 'scaling-law'~\cite{kaplan2020scaling, hoffmann2022training}. For some crucial abilities, such as reasoning, scaling the model has gradually transformed these abilities from a very low state to a usable state, and even approaching human capabilities. In this section, we provide an overview of the usage of LLMs in terms of the abilities and behaviors of LLMs along with scaling.

\begin{applebox}{Remark 5}
\begin{enumerate}[leftmargin=0.4cm]
\item With the exponential increase of model scales, LLMs become especially capable of reasoning like arithmetic reasoning and commonsense reasoning.
\item Emergent abilities become serendipity for uses that arise as LLMs scale up, such as ability in word manipulation and logical ability.
\item In many cases, performance does not steadily improve with scaling due to the limited understanding of how large language models' abilities change as they scale up.  
\end{enumerate}
\end{applebox}

\subsubsection{Use Case with Reasoning} Reasoning, which involves making sense of information, drawing inferences, and making decisions, is one of the essential aspects of human intelligence. It is challenging for NLP. Many existing reasoning tasks can be classified into commonsense reasoning and arithmetic reasoning.

\noindent\textbf{Arithmetic reasoning/problem solving}. 
The arithmetic reasoning capability of LLMs benefits greatly from the scaling of model size. For GPT-3, the ability of two-digit addition only becomes apparent when the number of parameters exceeds 13B~\cite{brown2020language}. Tasks to test arithmetic reasoning are trivial for humans 
and designed to challenge the capability of transferring natural language into mathematical symbols and multi-step inference. On GSM8k~\cite{cobbe2021training}, SVAMP~\cite{patel2021nlp} and AQuA~\cite{ling2017program}, LLMs, as generalists, have competitive performance with most methods which have task-specific designs. And GPT-4 overperforms any other methods~\cite{openai2023gpt4}, even some huge models particularly tuned for arithmetic problems~\cite{uesato2022solving}. Nevertheless, it should be noted that, without the intervention of external tools, LLMs may occasionally make mistakes in performing basic calculations, although chain-of-thought~(CoT) prompting~\cite{wei2022chain} can significantly improve LLMs' ability in calculations.

\noindent\textbf{Commonsense reasoning}. Commonsense reasoning not only requires LLMs to remember factual knowledge but also requires LLMs to do several inference steps about the facts. Commonsense reasoning increases gradually with the growth of model size. Compared to fine-tuned models, LLMs keep the superiority on most datasets, such as StrategyQA~\cite{geva2021did} and ARC-C~\cite{clark2018think}. Especially on  ARC-C, which contains difficult questions in science exams from grade 3 to grade 9, GPT-4 has been close to the performance of 100\% ~(96.3\%)~\cite{openai2023gpt4}.

\subsubsection{Use Cases with Emergent Abilities}Scaling of models also endows the model with some unprecedented, fantastic abilities that go beyond the power-law rule. These abilities are called "emergent ability". As defined in \cite{wei2022emergent}, \textit{emergent abilities of LLMs are abilities that are not present in smaller-scale models but are present in large-scale models}. This means such abilities cannot be predicted by extrapolating the performance improvements on smaller-scale models and the model suddenly gains good performance on some tasks once the scale exceeds a certain range. The emergent ability is typically unpredictable and surprising, leading to tasks that emerge randomly or unexpectedly. We examine concrete examples of the emergent abilities of LLMs and provide them as an important reference for deciding whether to leverage LLMs' emergent abilities.

Handling word manipulation is a typical emergent ability. It refers to the ability to learn symbolic manipulations, such as the reversed words~\cite{brown2020language}, in which the model is given a word spelled backwards, and must output the original word. For example. GPT-3~\cite{brown2020language} shows the emergent ability for word sorting, and word unscrambling tasks. PaLM~\cite{chowdhery2022palm} exhibits the emergent ability on ASCII word recognition~\footnote{Asking models to identify the word displayed as ASCII art, https://github.com/google/BIG-bench/tree/main/bigbench/benchmark\_tasks/ascii\_word\_recognition} and hyperbaton~\footnote{Asking models to choose the English sentence with adjectives in the "correct" order within two choices, https://github.com/google/BIG-bench/tree/main/bigbench/benchmark\_tasks/hyperbaton} task. 
The logical abilities of language models tend to emerge as the model scales up, such as logical deduction, logical sequence, and logic grid puzzles. Additionally, other tasks, such as advanced coding (e.g., auto debugging, code line description), and concept understanding (e.g., novel concepts, simple Turing concepts), are also use cases with the emergent abilities of large language models.

\subsubsection{No-Use Cases and Understanding} Although in most cases, as discussed above, larger models bring better performance, there are still many exceptions that should be considered when choosing the appropriate model. 

On certain tasks, with the size of LLMs increasing, the performance begins to decrease, such as Redefine-math: tests whether language models are able to work with common symbols when they are redefined to mean something else; Into-the-unknown: requires the model to choose which piece of information would help answer a question; Memo-trap: asks an LM to write a phrase in a way that starts like a famous quote but ends differently\footnote{More such tasks include: modus-tollens, pattern-matching-suppression, prompt-injection, repetitive-algebra and sig-figs. You can check them on: https://github.com/inverse-scaling/prize}. This is also called \textit{Inverse Scaling Phenomenon}.
Another interesting phenomenon observed in the scaling of LLMs is called the \textit{U-shaped Phenomenon}~\cite{wei2022inverse}. As the name implies, This phenomenon refers to that as LLM size increases, their performance on certain tasks initially improves but then starts to decline before eventually improving again, such as on: Hindsight-neglect: it tests whether language models are able to assess whether a bet was worth taking based on its expected value; NegationQA: this task takes an existing multiple-choice dataset and negates a part of each question to see if language models are sensitive to negation; Quote-repetition: it asks models to repeat back sentences given in the prompt, with few-shot examples to help it recognize the task. 
Hence the risk of diminishing performance should be noted and if the task is similar to those we just discussed, careful consideration should be given to whether or not to use huge LLMs. 

Gaining a deeper understanding of emergent abilities, inverse scaling phenomenon and U-shape phenomenon in LLMs is essential for advancing research in this field. In a certain sense, the U-shape phenomenon suggests that small-scale models and huge-scale models make predictions with different internal mechanisms. From this perspective, the U-shape phenomenon can be seen as a transformation of the inverse-scaling phenomenon due to some emergent abilities from sufficiently large models~\cite{wei2022inverse}. GPT-4~\cite{openai2023gpt4} exhibits a reversal of the inverse scaling phenomenon in some cases, such as on a task called Hindsight Neglect. The explanation for these behaviors of LLMs during scaling is still an open problem. Several hypotheses have been proposed. For emergent abilities, one explanation is that there may be multiple key steps for a task and the LLM cannot handle this task until it's large enough to handle every step, and another explanation is focused on the granularity of evaluation metrics~\cite{wei2022emergent}. For inverse-scaling phenomenon and u-shape phenomenon, the explanations mainly focus on the model's over-reliance on information from its prior rather than the input prompts, valid but misleading few-shot examples, and distracting easier tasks within a hard task~\cite{wei2022inverse}.

\subsection{Miscellaneous tasks}
This section explores miscellaneous tasks which cannot be involved in previous discussions, to better understand LLMs' strengths and weaknesses. 

\begin{applebox}{Remark 6}
\begin{enumerate}[leftmargin=0.4cm]
    \item Fine-tuned models or specified models still have their space in tasks that are far from LLMs' pretraining objectives and data.
    \item LLMs are excellent at mimicking human, data annotation and generation. They can also be used for quality evaluation in NLP tasks and have bonuses like interpretability.
\end{enumerate}
\end{applebox}

\subsubsection{No use case} LLMs generally struggle with some tasks due to differences in objectives and training data.

Although LLMs have achieved remarkable success in various natural language processing tasks, their performance in regression tasks has been less impressive. For example, ChatGPT's performance on the GLUE STS-B dataset, which is a regression task evaluating sentence similarity, is inferior to a fine-tuned RoBERTa performance \cite{zhong2023can}. The  Regression tasks typically involve predicting a continuous value rather than a discrete label, posing unique challenges for LLMs. One primary reason for their subpar performance is the inherent difference between the language modeling objective and the regression task objective. LLMs are designed to predict the next word in a sequence or generate coherent text, with their pre-training focused on capturing linguistic patterns and relationships. Consequently, their internal representations may not be well-suited for modeling continuous numerical outputs. 
Besides, LLMs have predominantly been trained on text data, focusing on capturing the intricacies of natural language processing. As a result, their performance on multimodal data, which involves handling multiple data types such as text, images, audio, video, actions, and robotics, remains largely unexplored. And fine-tuned multimodal models, like BE\textsc{i}T\cite{wang2022image} and PaLI~\cite{chen2022pali}, still dominate many tasks such as visual question answering~(VQA) and image captioning. Nonetheless, the recently introduced GPT-4~\cite{openai2023gpt4} has taken the step in multimodal fusion, but there is still a lack of detailed evaluation of its capabilities.

\subsubsection{Use case} LLMs are particularly suitable for certain tasks.

LLMs are very good at mimicking humans, acting as a chatbot, and performing various kinds of tasks. The LLMs-powered ChatGPT\footnote{https://chat.openai.com} is surprising for its consistency, reliability, informativeness, and robustness during multiple utterances with humans. The human-feedback procedure  plays an important role in acquiring such abilities

LLMs can both act as a good annotator and data generator for data augmentation, such as in\cite{yoo2021gpt3mix,dai2023chataug,ding2022gpt,tang2023does, yuan2023llm}. Some LLMs have been found as good as human annotators~\cite{gilardi2023chatgpt} in some tasks. And the collected texts from GPT-3.5~(text-davinci-003) have been used as human-like instruction-following demonstrations to train other language models~\cite{alpaca}. 

LLMs can also be used for quality assessment on some NLG tasks, such as summarization and translation. On summarization tasks, GPT-4 as an evaluator achieves a higher correlation with humans than other methods with a large margin~\cite{liu2023gpteval}. Some other evaluators based on LLMs~\cite{kocmi2023large, fu2023gptscore,liu2023gpteval,wang2023chatgpt} also show good human alignment in more NLG tasks, especially compared with traditional automatic metrics. But the LLM evaluator may have a bias towards the LLM-generated texts~\cite{liu2023gpteval}.

Also, as we discussed above, some abilities of LLMs bring bonuses in addition to performance improvement, such as interpretability. The CoT reasoning ability of LLMs can show how an LLM reaches the prediction, which is a good interpretation on the instance level, while it also improves the performance.

\subsection{Real world "tasks"}

In the last part of this section, we would like to discuss the usage of LLMs and fine-tuned models in real-world "tasks". We use the term "tasks" loosely, as real-world scenarios often lack well-formatted definitions like those found in academia. Many requests to models even cannot be treated as NLP tasks. Models face challenges in the real world from three perspectives:
\begin{itemize}
    \item \textbf{Noisy/Unstructured input}. Real-world input comes from real-world non-experts. They have little knowledge about how to interact with the model or even cannot use texts fluently. As a result, real-world input data can be messy, containing typos, colloquialisms, and mixed languages, unlike those well-formed data used for pre-training or fine-tuning. 
    \item \textbf{Tasks not formalized by academia}.In real-world scenarios, tasks are often ill-defined by academia and much more diverse than those in academic settings. Users frequently present queries or requests that do not fall neatly into predefined categories, and sometimes multiple tasks are in a single query. 
    \item \textbf{Following users' instructions}. A user's request may contain multiple implicit intents~(e.g. specific requirement to output format), or their desired predictions may be unclear without follow-up questions. Models need to understand user intents and provide outputs that align with those intents. 
\end{itemize}
Essentially, these challenges in the real world come from that users' requests deviate significantly from the distribution of any NLP datasets designed for specific tasks. Public NLP datasets are not reflective of how the models are used~\cite{ouyang2022training}. 
\begin{applebox}{Remark 7}
    LLMs are better suited to handle real-world scenarios compared to fine-tuned models. However, evaluating the effectiveness of models in the real world is still an open problem.
\end{applebox}

Handling such real-world scenarios requires coping with ambiguity, understanding context, and handling noisy input. Compared to fine-tuned models, LLMs are better equipped for this because they have been trained on diverse data sets that encompass various writing styles, languages, and domains. Additionally, LLMs demonstrate a strong ability to generate open-domain responses, making them well-suited for these scenarios. Fine-tuned models, on the other hand, are often tailored to specific, well-defined tasks and may struggle to adapt to new or unexpected user requests. They heavily rely on clear objectives and well-formed training data that specify the types of instructions the models should learn to follow. Fine-tuned models may struggle with noisy input due to their narrower focus on specific distributions and structured data. An additional system is often required as an assistant for fine-tuned models to process unstructured context, determine possible intents, and refine model responses accordingly.

Additionally, some mechanics such as instruction tuning~\cite{wei2021finetuned, sanh2021multitask} and human alignment tuning~\cite{ouyang2022training} further boost the capabilities of LLMs to better comprehend and follow user instructions. These methods improve the model's ability to generate helpful, harmless, and honest responses while maintaining coherence and consistency~\cite{wei2021finetuned, sanh2021multitask, ouyang2022training}. While both methods can make LLMs better generalize to unseen tasks and instructions, it has been noticed that while human labelers prefer models tuned for human alignment~\cite{ouyang2022training} to models tuned with instructions from public NLP tasks, such as FLAN~\cite{wei2021finetuned} and T0~\cite{sanh2021multitask}.  The reason may be similar to reasons for fine-tuned models' inferiority: public NLP tasks/datasets are designed for easy and automatic evaluation, and they can only cover a small part of real-world usage.

One of the main issues when it comes to real-world scenarios is how to evaluate whether the model is good or not. Without any formalized tasks or metrics, the evaluation of model effectiveness can only rely on feedback from human labelers. Considering the complexity and cost of human evaluation, there's no massive and systematic comparison between fine-tuned models and LLMs yet. Nevertheless, the huge success and popularity of LLMs such as chatGPT, have confirmed the superiority of LLMs to some extent.

\section{Other considerations}

Despite LLMs are suitable for various downstream tasks, there are some other factors to consider, such as efficiency and trustworthiness. 
Our discussion of efficiency encompasses the training cost, inference latency, and parameter-efficient tuning strategies for LLMs. Meanwhile, our examination of trustworthiness includes robustness \& calibration, fairness \& biases, potential spurious correlations, and the safety challenges in LLMs.

\begin{applebox}{Remark 8}
\begin{enumerate}[leftmargin=0.4cm]
    \item Light, local, fine-tuned models should be considered rather than LLMs, especially for those who are sensitive to the cost or have strict latency requirements. Parameter-Efficient tuning can be a viable option for model deployment and delivery.
    \item  The zero-shot approach of LLMs prohibits the learning of shortcuts from task-specific datasets, which is prevalent in fine-tuned models. Nevertheless, LLMs still demonstrate a degree of shortcut learning issues.
    \item  Safety concerns associated with LLMs should be given utmost importance as the potentially harmful or biased outputs, and hallucinations from LLMs can result in severe consequences. Some methods such as human feedback have shown promise in mitigating these problems.
\end{enumerate}
\end{applebox}

\subsection{Efficiency}
In real-world deployment, performance, cost, and latency are all important considerations, not just the performance of the models. While some parameter-efficient methods have been developed, practitioners must balance efficiency with effectiveness in the practice. 

\paragraph{Cost}
LLMs have grown increasingly larger in recent years, with models such as GPT-1, GPT-2, and GPT-3 featuring 117 million, 1.5 billion, and 175 billion parameters, respectively. The cost of training an LLM is heavily influenced by its size, with estimates suggesting that training the 11B parameter variant of T5 costs well over \$1.3 million for a single run, while a single training run of GPT-3 175B requires \$4.6 million~\cite{OpenAIsG17:online}.
The energy consumption for training large models is equally impressive. The total energy consumption for training a transformer model with 6B parameters to completion is estimated to be around 103.5 MWh~\cite{dodge2022measuring}. Google reports that training PaLM consumed about 3.4 GWh in about two months~\cite{ananthaswamy2023ai}.
Furthermore, the dataset size also scales rapidly with the size of the model, with GPT-3 175B trained on 499 billion tokens~\cite{brown2020language}. Another key metric that reflects the computing cost is Flops, with GPT-3 175B requiring $3.14 \times 10^{23}$ Flops, while a T5 11B model only requires $3.30 \times 10^{22}$, which is 10 times less.
In addition to these costs, hardware requirements are also substantial.
OpenAI has collaborated with Microsoft on a supercomputer hosted in the Microsoft Azure cloud, consisting of 285k CPU cores and 10k high-end GPUs to support the training of large models.
For users of the OpenAI API, pricing varies based on the model and usage, with options such as GPT-3.5-turbo charging \$0.002 per 1k tokens for chat service. However, for users who require custom models, training costs \$0.03 per 1k tokens, while usage costs \$0.12 per 1k tokens~\cite{Pricing84:online}.
Therefore, for users who cannot afford such a large cost, such as small startups, individual users, etc., a small, fine-tuned model is a better and more reasonable choice.

\paragraph{Latency}
Latency is a crucial factor to consider in real-world applications of LLMs. Inference time is a commonly used metric to measure latency, which is highly dependent on the model size, architecture, and token size. For instance, the inference time for the GPT-J 6B model is 0.077s, 0.203s, and 0.707s when the max token size is set to 2, 8, and 32, respectively. Additionally, when the max token size is fixed at 32, the inference time for the InstructGPT model~(davinci v2) is 1.969s.
As LLMs are often too large to be run on a single user's machine, companies provide LLM services via APIs. The API latency can vary depending on the user's location, and the average latency of the OpenAI API service for a single request can range from a few hundred milliseconds to several seconds.
In scenarios where high latency is not acceptable, large LLMs may not be appropriate. For example, scalability is critical in many information retrieval applications. To deploy information retrieval systems on the web, search engines require very efficient inference for systems to be useful. The idealized denoised inference time for the InstructGPT davinci v2 (175B*) model is 0.21s per request (i.e., a query-passage pair to be scored), which is too slow for web search engines.

\paragraph{Parameter-Efficient Tuning} In practice, we may tune the model on some specific datasets. Parameter-Efficient Tuning (PET) is an efficient technique to tune a small portation of model parameters (or extra parameters) while freezing most parameters of the pre-trained LLMs. The main goal of PEFT is to greatly decrease the computational and storage costs while keeping the performance of the original models. The common techniques for PET are LoRA~\cite{hu2021lora}, Prefix Tuning~\cite{li2021prefix}, P-Tuning~\cite{liu2022p,liu2021p2}. As an illustration, the LoRA method maintains the weights of the pre-trained model and incorporates low-rank matrices into every layer of the Transformer architecture. This approach considerably minimizes the number of parameters that require training for subsequent tasks, thereby increasing overall efficiency. Alpaca-LoRA\footnote{https://github.com/tloen/alpaca-lora} proposes integrating Low-Rank Adaptation (LoRA) into LLaMA-Alpaca, which enables runs LLaMA within hours on a single RTX 4090. All these PFT methods can be helpful either for fine-tuning a model to a specific task or tuning LLMs to meet special requirements like human alignment.

\subsection{Trustworthiness}
Given that LLMs are now involved in sensitive areas such as healthcare, finance, and law, it is crucial to ensure that they are trustworthy and capable of producing reliable output.

\paragraph{Robustness and Calibration}
The accuracy and robustness of the LLMs are shown to have a very strong correlation~\cite{liang2022holistic}. The models that have high accuracy on the scenario also have good robustness. However, the robustness of the zero-shot becomes worse after being tuned on extra application-specific tasks data~\cite{wortsman2022robust}. This may due to overfitting, which leads to poor generalizability due to the extremely high complexity of the model and the limited training samples from downstream tasks~\cite{hua2022fine}.
In a similar vein, it has been observed that fine-tuning a model can result in significant miscalibrations, owing to over-parameterization~\cite{kong2020calibrated}. Therefore, fine-tuned models may not be an optimal choice when robustness and calibration are critical considerations.
However, human alignment has been found as a potential solution for enhancing model robustness. InstructGPT davinci v2 (175B*) has been shown to outperform other models in terms of robustness. On the other hand, achieving optimal calibration of the model depends on the scenario and adaptation procedure employed.

\paragraph{Fairness and Bias}
LLMs have been shown to exhibit disparate treatment and impact, perpetuating societal biases and potentially leading to discrimination~\cite{buolamwini2018gender, beukeboom2019stereotypes}. To ensure fairness and equity for all users, it is crucial to address these issues in the development and deployment of NLP models. Disparities in performance between demographic groups can serve as an indicator of fairness problems.
LLMs are particularly susceptible to fairness issues, as significant performance disparities have been observed across demographic categories such as dialect, religion, gender, and race~\cite{liang2022holistic}. However, research has shown that aligning models with human instructions can improve LLM performance regardless of their size, with the InstructGPTmodel~(davinci v2)  exhibiting smaller performance disparities than other LLMs~\cite{chung2022scaling}. 

\paragraph{Spurious Biases}
The shortcut learning problem has been observed in various natural language understanding tasks under the pretraining and fine-tuning paradigm, where models heavily rely on spurious correlations between input and labels in the fine-tuning data for prediction \cite{geirhos2020shortcut, tang2021mitigating, du2022shortcut}. For example, in reading comprehension tasks, fine-tuned models tend to focus on the lexical matching of words between the question and the original passage, neglecting the intended reading comprehension task itself \cite{lai2021machine}. %
In contrast, large language models are not directly trained on fine-tuned datasets, which makes it less likely for them to learn shortcut features present in the fine-tuned dataset, thereby enhancing the model's generalization capabilities. However, LLMs are not infallible and may exhibit some shortcut learning during in-context learning. For example, recent preliminary studies have begun investigating the robustness of prompt-based methods in large-scale language models \cite{zhao2021calibrate, webson2022prompt}. One such study evaluates the few-shot learning performance of GPT-3 on text classification and information extraction tasks \cite{zhao2021calibrate}. and reveal that the examined LLMs are susceptible to majority label bias and position bias, where they tend to predict answers based on the frequency or position of the answers in the training data. Moreover, these LLMs exhibit common token bias, favoring answers that are prevalent in their pre-training corpus. Recent studies show that this positional bias can be mitigated by selecting proper prompts \cite{lu2022fantastically}.
In summary, while LLMs significantly reduce the shortcut learning problem prevalent in fine-tuned models, they still exhibit some shortcut learning issues and should be approached with caution when deploying them in downstream applications.

\subsection{Safety challenges}
LLMs have demonstrated their extremely strong capabilities in many areas such as reasoning, knowledge retention, and coding. As they become more powerful and human-like, their potential to influence people's opinions and actions in significant ways grows. As a result, some new safety challenges to our society should be considered and have caught lots of attention in recent works~\cite{openai2023gpt4, openai2023gpt4-sys}.

\paragraph{Hallucinations} 

The potential for LLMs to "hallucinate," or generate nonsensical or untruthful content, can have significant negative impacts on the quality and reliability of information in various applications. As LLMs become increasingly convincing and believable, users may develop an overreliance on them and trust them to provide accurate information in areas with which they are somewhat familiar. This can be particularly dangerous if the model produces content that is entirely false or misleading, leading to incorrect decisions or actions taken based on that information. Such outcomes can have serious consequences in many domains, such as healthcare, finance, or public policy, where the accuracy and reliability of information are critical. To mitigate these issues, reinforcement learning from human feedback (RLHF) is widely used~\cite{ouyang2022training, openai2023gpt4-sys}  and LLMs themselves have been integrated into the loop~\cite{openai2023gpt4-sys}.

\paragraph{Harmful content}
Due to the high coherence, quality, and plausibility of texts generated by LLMs, harmful contents from LLMs can cause significant harm, including hate speech, discrimination, incitement to violence, false narratives, and even social engineering attack. The implementation of safeguards to detect and correct those contents can be mitigation~\cite{tang2023science}. These LLMs can also have dual-use potential by providing required illicit information, leading to risks such as the proliferation of weapons~\cite{openai2023gpt4-sys} and even terrorism attack planning. It is crucial to ensure using these LLMs responsibly, with safeguards in place to prevent harm. Also, in existing work, feedback from humans plays an important role in getting rid of harmful outputs. 

\paragraph{Privacy} LLMs can face serious security issues. An example is the issue of user privacy. It is reported that Samsung employees were using ChatGPT to process their work when they inadvertently leaked top-secret data, including the source code proper of the new program, internal meeting minutes related to the hardware, etc. The Italian data protection agency declared that OpenAI, the developer of ChatGPT, illicitly gathered personal user data, leading Italy to become the first government to prohibit ChatGPT over privacy concerns~\cite{ChatGPTI90:online}.

\section{Conclusion and Future Challenges}

Recent advances in large language models have been revolutionizing the field of natural language processing. Effectively using LLMs requires understanding their capabilities, and limitations for various NLP tasks. This work presents a practical guide to working with LLMs for downstream NLP tasks. We first discuss prominent models like GPT-style and BERT-style architectures and the factors influencing their performance. We then explore using LLMs for downstream tasks, including knowledge-intensive tasks, NLU, and NLG tasks, as well as providing concrete examples of successes and limitations. This practical guide offers insights into LLMs and best practices for harnessing LLMs across NLP tasks. We hope it would enable researchers and practitioners to leverage their potential and drive innovation in language technologies.

In the following, we figure out the future challenges of the LLMs:
\begin{itemize}[leftmargin=0.4cm]
    \item\textbf{Evaluation of proposed models on real-world ``datasets''.} 
    While existing deep learning models are primarily evaluated on standard academic datasets, such as ImageNet, which have been milestones in deep learning development. However, the limitations of standard academic datasets can not exactly reflect real-world performance. As models advance, it is crucial to assess them on more diverse, complex, and realistic data that reflect real-world needs. Evaluating models on real-world ``datasets'', in addition to academic ones, will provide a more rigorous test of their capabilities, as well as a better understanding of their effectiveness in real-world applications. This ensures that the models are capable of addressing real-world challenges and delivering practical solutions. 
    
    \item \textbf{Model Alignment.} 
    Ensuring that increasingly powerful and autonomous models align with human values and priorities is essential. Methods must be developed to guarantee that these models behave as intended and do not optimize for undesirable outcomes. It is crucial to integrate alignment techniques from the start of the model development process. Model transparency and interpretability are also important factors for evaluating and ensuring alignment. Additionally, as we look toward the future, an even more daunting challenge looms: aligning superhuman systems. While this task is currently beyond our demands, it is important to consider and prepare for the potential implications of aligning such advanced systems, as they may present unique complexities and ethical concerns~\cite{bai2022constitutional, bowman2022measuring}. 
        
    \item\textbf{Safety Alignment.} While discussion of AI existential risks is important, concrete research is needed to guarantee the safe development of advanced AI. This includes techniques for interpretability, scalable oversight and governance, and formal verification of model properties. Safety should be considered not just an add-on but an integral part of the model-building process.
    
    \item\textbf{Performance Prediction with Scaling.} It is difficult to anticipate how model performance will change as model size and complexity increases dramatically. Developing methods to better predict model performance after scaling up or as new architectures are developed would allow for more efficient use of resources and accelerated progress. Some possibilities include: training a smaller 'seed' model and extrapolating its growth, simulating the effects of increased scale or model tweaks, and benchmarking iterations of the model at different scales to build scaling laws. These could provide insight into the performance of models even before they are built.
\end{itemize}

\bibliographystyle{plain}
\bibliography{./H.J, ./Q.F, ./X.H, ./R.X}

\end{document}